%% file: acl_latex.tex
\pdfoutput=1

\documentclass[11pt]{article}

\usepackage[]{acl}

\usepackage{times}
\usepackage{latexsym}
\usepackage{booktabs}
\usepackage{tikz}
\usepackage{pgfplots}
\usepackage{caption}
\usepackage{subcaption}
\pgfplotsset{compat=1.18}
\usepackage{array}

\usepackage[T1]{fontenc}

\usepackage[utf8]{inputenc}

\usepackage{microtype}

\usepackage{inconsolata}

\usepackage{graphicx}

\title{Scaling Multi-Document Event Summarization: Evaluating Compression vs. Full-Text Approaches}

\author{Adithya Pratapa \qquad Teruko Mitamura \\
    Language Technologies Institute \\
    Carnegie Mellon University \\
    \texttt{\{vpratapa, teruko\}@cs.cmu.edu} \\
}

\begin{document}
\maketitle
\begin{abstract}
Automatically summarizing large text collections is a valuable tool for document research, with applications in journalism, academic research, legal work, and many other fields. In this work, we contrast two classes of systems for large-scale multi-document summarization (MDS): compression and full-text. Compression-based methods use a multi-stage pipeline and often lead to lossy summaries. Full-text methods promise a lossless summary by relying on recent advances in long-context reasoning. To understand their utility on large-scale MDS, we evaluated them on three datasets, each containing approximately one hundred documents per summary. Our experiments cover a diverse set of long-context transformers (Llama-3.1, Command-R, Jamba-1.5-Mini) and compression methods (retrieval-augmented, hierarchical, incremental). Overall, we find that full-text and retrieval methods perform the best in most settings. With further analysis into the salient information retention patterns, we show that compression-based methods show strong promise at intermediate stages, even outperforming full-context. However, they suffer information loss due to their multi-stage pipeline and lack of global context. Our results highlight the need to develop hybrid approaches that combine compression and full-text approaches for optimal performance on large-scale multi-document summarization.\footnote{Our code and data are available at \url{https://github.com/adithya7/scaling-mds}.}
\end{abstract}

\section{Introduction}

Summarizing events described in document collections has long interested the NLP community with shared tasks for event tracking \cite{allan-etal-1998-tdt} and summarization \cite{chieu-etal-2004-query, dang-owczarzak-etal-2009-overview, aslam-etal-2015-trec}. Given an input collection of hundreds of text documents, systems have to extract and summarize salient information about the event. The length and diversity of the input presents a challenge to recent large language models (LLMs). In this work, we contrast two classes of systems for large-scale multi-document summarization (MDS), compression-based, and full-text systems.\footnote{We use the term \emph{scale} to refer to the large number of documents associated with each summary.}

Full-text systems promise a lossless approach by providing the summarizer access to the entire input. They are based on the long-context reasoning abilities of LMs, having already shown strong retrieval performance on long inputs \cite{hsieh-etal-2024-ruler}. However, their capabilities on large-scale MDS are not as well understood. In a recent work, \citet{laban-etal-2024-summhay} introduced a synthetic MDS benchmark that resembles the Needle in a Haystack evaluation \cite{kamradt-etal-2023-needle}. In addition to this dataset, we evaluate on two large-scale event summarization datasets: Background \cite{pratapa-etal-2023-background} and WCEP \cite{gholipour-ghalandari-etal-2020-large}. We contrast the end-to-end full-context method\footnote{We use full-text and full-context interchangeably.} with three compression-based methods: retrieval, hierarchical, and incremental. Each method \emph{compresses} the input in a multistage pipeline (\S\ref{ssec:methods}). We evaluated the content selection aspects of the summary using the Atomic Content Unit (A3CU) metric \cite{liu-etal-2023-towards-interpretable}.

Our experiments show that full-context and retrieval perform best in most settings (\S\ref{sec:results}). To better understand the performance of compression-based methods, we measure A3CU recall to track the salient information retention in their intermediate outputs (\S\ref{ssec:full_context_vs_compression}). Across all settings, we find that compression-based methods show high recall in intermediate stages but suffer information loss in their multistage pipeline. In particular, the intermediate recall is often much higher than the full-context system recall. We highlight two key takeaways: First, while iterative methods (hierarchical \& incremental) were previously found effective for book summarization and small-scale MDS, they underperform on large-scale MDS. Second, full-context systems are suboptimal on large-scale MDS datasets. We advocate for hybrid methods that combine input compression and long-context models. Such hybrid approaches are also scalable to even larger MDS tasks that go far beyond the context window limits of current LLMs.

\section{Experimental Setup}

\subsection{Datasets}
\label{ssec:datasets}

Our three datasets provide different flavors of the multi-document summarization task (\autoref{tab:datasets}).

\textbf{SummHay:} A query-focused dataset that covers the news and conversation domains \cite{laban-etal-2024-summhay}. Synthetically generated using GPT-3.5 and GPT-4o, each summary constitutes a set of insights. To keep our evaluation setup consistent across datasets, we concatenate these insights into a free-form summary. Following the original work, we include an oracle setting that only retains documents containing the reference insights.

\textbf{Background:} This dataset provides summaries of complex news events \cite{pratapa-etal-2023-background}. The task is based on an event timeline. For a given day, the goal is to generate a background summary by summarizing past new articles related to the event. We expand the original dataset to use news articles instead of just news updates. The dataset includes three human-written background summaries.

\textbf{WCEP:} A newswire dataset collected from Wikipedia Current Events Portal \cite{gholipour-ghalandari-etal-2020-large}. The summaries come from the portal and the documents include a combination of cited source articles and a retrieved collection of related articles from the Common Crawl archive.

Our choice of datasets collectively represents the real-world use-cases of multi-document summarization systems. Previous work has shown the effectiveness of full-context methods in retrieval tasks. To this end, we include the query-focused SummHay dataset. On the other hand, Background and WCEP provide different variants of the task. Background task requires accumulation of salient content units over the entire input. WCEP has high information redundancy, with many articles providing support for the salient units.

\input{tables/datasets}

\subsection{Methods}
\label{ssec:methods}

\input{tables/scores_all}

We now describe our long-context methods and transformers. The key difference between our methods is the length of the input passed to the summarization system (transformer) at any stage.

\textbf{Full-context:} The transformer has access to the full input and relies on its long context reasoning abilities to generate the summary.

\textbf{Iterative:} Multi-stage summarization where we iteratively pass chunks of the input to the transformer. We explore two methods, hierarchical and incremental. The hierarchical method summarizes each document and iteratively merges these to compile the final summary. The incremental method processes documents in order while maintaining a running summary of the input. Previous work explored these methods for book summarization \cite{chang-etal-2024-booookscore} and small-scale multi-document summarization \cite{ravaut-etal-2024-context}.

\textbf{Retrieval:} We rank the input documents according to their relevance to the query.\footnote{If a query is unavailable, we default to using `Generate a summary of the document' as the query.} We then select the top-ranked documents (up to 32k tokens) and pass their concatenation to the transformer. We use SFR Embedding-2 \cite{meng-etal-2024-sfr} for the retrieval task and order-preserving RAG following the recommendation from \citet{yu-etal-2024-defense-rag}. We set 32k as the limit because all of our transformers are effective at this context length \cite{hsieh-etal-2024-ruler}.

\subsection{Transformers}
\label{ssec:transformers}

For our summarization systems, we experiment with three transformer-based models, Llama-3.1, Command-R, and Jamba-1.5. Each model supports a context window of at least 128k tokens. They rely on a different long-context methodologies, and represent the broad class of open-weight LLMs. All the three models show competitive performance on the RULER benchmark for long-context LMs \cite{hsieh-etal-2024-ruler}.

\textbf{Llama-3.1:} Pretrained on 15T+ tokens, it supports long context by using a large base frequency of 500,000 and non-uniform scaling of RoPE dimensions \cite{llama31-modelcard}. We use both 8B and 70B variants to test the effect of model scaling.

\textbf{Command-R:} A transformer-based model that uses NTK-aware interpolation with a very large RoPE base frequency of 4M \cite{cohere-for-ai-2024-commandr}. We use the 32B variant.

\textbf{Jamba-1.5:} A hybrid architecture with interleaved Transformer and Mamba layers \cite{jambateam-2024-jamba15}. It involves both mid-training on long texts and post-training on (synthetic) long-context tasks. We use the 52B Jamba-1.5-Mini mixture-of-experts model with 12B active parameters.

For a fair comparison of above methods and transformers, we set the maximum input length to 128k across all settings. If the input is longer than 128k tokens, we first truncate the longest documents. In the case of Background, we also ensure equal representation from the past events by budgeting the token limit to each past timestamp. We also set a minimum document length (128 tokens) and drop documents if this cannot be achieved. To ensure that all methods see the same input, we adopt the same truncation strategy across full-text and compression-based methods. Theoretically, compression-based methods could work with even longer input (>128k), but we limit all settings to 128k tokens for a fair comparison.

See \S\ref{app:experimental_setup} in the Appendix for additional details about our experimental setup including our summarization prompt (\autoref{tab:summarization_prompt}). We sample summaries with a temperature of 0.5. We note that the summaries could be slightly different across different seeds. \citet{vig-etal-2022-exploring} compared end-to-end and RAG for query-focused summarization, but limited to the short input setting.

\section{Results}
\label{sec:results}

\input{figures/inf_loss}

\subsection{Metrics}
\label{ssec:metrics}

We focus our analysis on the \emph{content selection} aspect of summarization. \citet{nenkova-passonneau-2004-evaluating} first studied the content selection evaluation using the pyramid method on summarization of content units. Follow-up efforts have automated various parts of this method \cite{shapira-etal-2019-crowdsourcing, liu-etal-2023-towards-interpretable}. In this work, we use the reference-based Atomic Content Unit (A3CU) metric \cite{liu-etal-2023-towards-interpretable} that is based on the definition of atomic content units of \citet{liu-etal-2023-revisiting}. This metric is trained to predict a score that measures the overlap of atomic content units between the reference and predicted summaries.

Recent works also studied faithfulness \cite{kim-etal-2024-fables}, coherence \cite{chang-etal-2024-booookscore}, and position bias \cite{huang-etal-2024-embrace,ravaut-etal-2024-context,laban-etal-2024-summhay}. Although these evaluations are important, content selection remains a core issue for large-scale MDS.

\subsection{Overall Results}
\label{ssec:overall_a3cu}

\autoref{tab:overall_a3cu} reports the A3CU F1 scores for compression-based methods relative to the full-context baseline.\footnote{We report ROUGE and A3CU precision, recall in \S\ref{app:full_metrics}.} Full-context and retrieval perform the best, being particularly effective on the query-focused SummHay dataset. The two iterative methods perform poorly in most settings. We also find that the performance of transformers and methods varies considerably across the datasets and even within examples in each dataset.\footnote{See \autoref{fig:box_a3cu} in the Appendix for example-level trends.} Below, we break down these results and analyze the effect of transformer and compression methods.

Due to the high costs of running API-based models on long texts, we mostly limit our evaluation to open-weight LLMs. We report preliminary results using Gemini-1.5 on SummHay in \autoref{tab:summhay_gemini} in the Appendix. We noticed trends similar to those of open-weight LLMs.

\subsection{Analysis: Full-context \& Transformer}

In the full-context setting, we see mixed results across transformers, with none performing the best across all datasets. Interestingly, Llama-3.1-8B outperforms 70B on SummHay. This surprising result aligns with their relative performance on the RULER benchmark at 128k context length. The 70B model fares better in the oracle setting and shows similar performance on non-retrieval-style datasets. We believe that the 70B model needs additional post-training to improve its long-context retrieval performance.

Command-R underperforms the much smaller Llama-3.1-8B. This could be attributed to its use of RoPE \cite{su-etal-2021-roformer}. Command-R increases the base frequency while Llama-3.1 additionally scales RoPE dimensions non-uniformly, likely leading to better long-context capabilities \cite{ding-etal-2024-longrope}. However, without specific details on the mid- and post-training with long texts, it would be difficult to identify the exact cause. We direct the reader to \citet{peng-etal-2023-yarn} and \citet{lu-etal-2024-controlled} for a discussion on long-context methods.

\subsection{Analysis: Full-context vs. Compression}
\label{ssec:full_context_vs_compression}

With the exception of retrieval on query-focused SummHay dataset, compression-based methods generally underperform full-context (\autoref{tab:overall_a3cu}). To analyze this, we use A3CU \emph{recall} to track the retention of salient information in intermediate outputs. These intermediate outputs correspond to the retrieved documents (retrieval) and intermediate summaries (hierarchical, incremental). \autoref{fig:inf_loss} reports the recall scores for the final summary and the best intermediate output (excl. final). For comparison, we also report the recall score for the full-context summary. Across datasets, the best intermediate recall is significantly higher than the final summary recall, even outperforming full-context.\footnote{Since recall is impacted by the summary length, we report average length of summaries for each system in \autoref{tab:summary_len_stats} in the Appendix. We do not find any noticeable correlation.}

We highlight two key observations. First, iterative methods suffer catastrophic information loss in their multistage pipeline. Second, the best intermediate recall scores from compression methods show areas of improvement for full-context systems. As a control setting, we evaluated on SummHay-oracle and found full-context to be comparable to the best intermediate recall from compression methods (\autoref{fig:inf_loss_summhayoracle} in the Appendix).

\paragraph{Retrieval:} Relative performance of full-context and retrieval varies widely across examples and transformers. \citet{karpinska-etal-2024-nocha} observed similar behavior for claim verification on books. In particular, for Llama-3.1-8B on SummHay, we find the final summary to be better than the best intermediate output (\autoref{fig:inf_loss}). This is the optimal scenario, illustrating the system's effectiveness in aggregating information from the retrieved documents. We do not see this behavior in other settings.

\paragraph{Iterative:} We qualitatively analyze the outputs from iterative methods. The hierarchical method tends to generate increasingly abstract summaries at higher levels. It often skips details such as entities and numerals in the summaries. We observe this behavior across all transformers. With the incremental method, we attribute poor performance to the large number of intermediate steps (\# documents). Even though the system retrieves salient information at an intermediate stage, the model often gets distracted by non-salient information seen in documents thereafter. We provide examples in \autoref{tab:qual_hierarchical} and \autoref{tab:qual_incremental} in the Appendix.

In the Appendix (\S\ref{ssec:ablations}), we also experiment with short-context transformers such as Llama-3 (\autoref{tab:summhay_llama3}), varying chunk sizes for the hierarchical method, an alternative embedding method for retrieval (\autoref{tab:summhay_e5}), and grounded generation templates for Jamba and Command-R.

\subsection{Human Evaluation}

To complement our automatic evaluation, we perform a reference-based human evaluation. We randomly sample 62 examples from the SummHay dataset ($\approx$67\%) and ask a human expert\footnote{This task was done by the first author.} to rate the system summaries. We follow recommendations from prior work \cite{kiritchenko-mohammad-2017-best, goyal-etal-2022-news, pratapa-etal-2023-background} to use the best-worst rating scale. For each example, the human evaluator picks the best and worst summaries (multiple allowed) among the four methods, full context, hierarchical, incremental, and retrieval (Llama-3.1-8B). They use reference summaries to perform content selection evaluation. We shuffle the presentation order of the system summaries in each example, and system labels are completely hidden from the human evaluator. The results of our human evaluation are presented in \autoref{tab:human_eval}. Retrieval-based summaries are rated the best, followed by full-context, incremental, and hierarchical. These results strongly correlate with our automatic evaluation (\autoref{tab:overall_a3cu}).

\input{tables/human_eval}

\subsection{Recommendations for Future Work}
\label{ssec:recommendations}

Based on our analysis, we make two recommendations for future work on large-scale MDS. First, hybrid systems that combine input compression methods with long-context LLMs. Second, a reference-free content selection evaluation that facilitates further scaling of MDS.

\paragraph{Hybrid Methods:} Our analysis using A3CU recall shows the scope for improvement of full-context systems (\autoref{fig:inf_loss}). Recent studies have shown that long-context models are not as effective as claimed for retrieval tasks \cite{hsieh-etal-2024-ruler,karpinska-etal-2024-nocha}, and our results support this for large-scale MDS. Iterative methods were previously used for book summarization \cite{chang-etal-2024-booookscore} and small-scale MDS \cite{ravaut-etal-2024-context}. In large-scale MDS, they show a significant loss of salient information. Based on these observations, we advocate for a hybrid approach that utilizes selective input compression methods \cite{sarthi-etal-2024-raptor,xu-etal-2024-recomp,jiang-etal-2024-longllmlingua} in conjunction with a long-context LLM. A hybrid approach could provide optimal performance while improving the runtime over full-context. It also allows for scaling to a very large-scale MDS that goes far beyond the model context window.

\paragraph{Reference-free evaluation:} In our analysis, we used a reference-based A3CU metric. As we scale the MDS task to include hundreds or thousands of documents, obtaining high-quality human-written reference summaries will be infeasible. Therefore, reference-free content selection evaluation metrics are needed. Synthetic tasks such as SummHay present a promising alternative.

\section{Conclusion}

In this work, we contrast the full-context method against three compression-based methods for large-scale MDS. We evaluated on three datasets, SummHay, Background, and WCEP using the A3CU content selection evaluation metric. We find that the full-context and retrieval-based methods perform the best. Iterative methods suffer from significant information loss. Our analysis shows that full-context methods provide suboptimal performance, and we recommend future work to explore hybrid methods that combine the strengths of input compression methods with advances in long-context LLMs.

\section*{Limitations}

In this work, we rely on high-quality reference summaries to measure the content selection aspects of system-generated summaries. We acknowledge that human evaluation is the gold standard for text summarization. However, for large-scale multi-document summarization ($\approx$100 docs per example), it is prohibitively expensive to perform human evaluation. \citet{karpinska-etal-2024-nocha} reported that a human takes about 8-10 hours to read an average book (of similar length to our setting). We leave the extension of human evaluation of full-context and compression-based systems to future work. We also limit our evaluation to models with publicly available weights. We report preliminary results on SummHay using Gemini-1.5 (\autoref{tab:summhay_gemini} in Appendix). Due to the high API costs of running Gemini on long inputs, we couldn't run them for other datasets. We did not conduct an extensive search for optimal prompts for the summarization task. So, it is possible that the performance of some system configurations could be improved with additional prompt tuning.

\section*{Ethics Statement}

Hallucination is an important concern for text summarization systems and has been widely studied in the literature. We focus on the content selection aspects of text summarization and choose our evaluation metrics accordingly. However, we recognize the importance of faithfulness evaluation in providing a holistic evaluation of summarization systems. We leave this extension to future work.

\section*{Acknowledgments}

We thank the ARR reviewers for their valuable feedback in improving our paper. Adithya Pratapa was supported by a LTI Ph.D. fellowship.

\bibliography{custom}

\appendix

\section{Appendix}
\label{sec:appendix}

We use GitHub copilot and Claude-3.5 Sonnet for assistance with coding and editing.

\subsection{Datasets}
\label{app:datasets}

For background summarization, we use the news articles from the original timeline summarization datasets, Timeline17 \cite{tran-etal-2013-predicting}, Crisis \cite{tran-etal-2015-timeline} and Social Timeline \cite{wang-etal-2015-socially}. To constrain the input length, we use a maximum of five news articles from any given day. We also experimented with prefiltering the articles using the news update of the given day, but this did not show improvements in summary quality.

\subsection{Experimental Setup}
\label{app:experimental_setup}

\textbf{Transformers:} We use weights from Huggingface for Llama-3.1-8B,\footnote{\url{https://hf.co/meta-llama/Llama-3.1-8B-Instruct}} Llama-3.1-70B,\footnote{\url{https://hf.co/meta-llama/Llama-3.1-70B-Instruct}} Command-R,\footnote{\url{https://hf.co/CohereForAI/c4ai-command-r-08-2024}} and Jamba-1.5-Mini.\footnote{\url{https://hf.co/ai21labs/AI21-Jamba-1.5-Mini}}

\textbf{Compute:} We run inference using vLLM on four 48G GPUs \cite{kwon-etal-2023-efficient}. Given its large size, we load Llama-3.1-70B with fp8 precision. For the smaller Llama-3.1-8B, we use a single 48G GPU. Our setup includes a mix of Nvidia's A6000, L40, and 6000 Ada GPUs.

\textbf{Iterative methods:} For both iterative methods, we set the maximum chunk size to 4096 tokens. For the hierarchical method, we first generate summaries for each input document. Then, we pack consecutive document summaries into the maximum chunk size for the next summarization step. We stop the process when we only have one summary. For the incremental method, we start by generating the summary of the first document. Then, we concatenate this summary with the following document for the next summarization step. We iterate through every document in the input, in the order provided by the dataset. The document order is relevant for Background (event timelines), but might not be as relevant for SummHay and WCEP.

\textbf{Retrieval:} We limit each document to 1024 tokens and the post-retrieval input to 32k tokens.

\textbf{Summary length:} To set the maximum summary words for each dataset, we first tokenize the summaries in the validation split using NLTK. We use the 80th percentile as the maximum summary words for the systems. To account for the differences in tokenizers for Llama-3.1, Command-R, and Jamba-1.5, we set the maximum number of summary \emph{tokens} by multiplying the maximum summary words with model-specific word-to-token ratios. The word-to-token ratios for Llama-3.1, Command-R, and Jamba-1.5-Mini are 1.145, 1.167, and 1.219 respectively. For iterative methods, we use the same maximum summary token limit at each intermediate step. In \autoref{tab:summary_len_stats}, we report the average length of system-generated summaries.

\textbf{Prompt:} \autoref{tab:summarization_prompt} provides our prompt for the text summarization task. We use the same prompt for all transformers and methods. We follow the recommendations from model providers and use the model-specific chat templates from Huggingface tokenizers when prompting the instruction-fine-tuned models.

\input{tables/prompts}
\input{figures/inf_loss_appendix}

\subsection{Full Metrics}
\label{app:full_metrics}

We report the precision, recall, and F1 scores for A3CU and ROUGE scores \cite{lin-2004-rouge} for each dataset: SummHay (\autoref{tab:summhay}), SummHay oracle (\autoref{tab:summhay_oracle}), Background (\autoref{tab:background}), and WCEP (\autoref{tab:wcep}). We use Huggingface evaluate for ROUGE and the original repo for A3CU.\footnote{\url{https://github.com/Yale-LILY/AutoACU}}

\subsection{Example-level Trends}

\autoref{fig:box_a3cu} shows the distribution of A3CU F1 scores across examples. We notice a significant variance in system performance across all datasets.

\input{figures/box_a3cu}

\subsection{Ablations}
\label{ssec:ablations}

We perform ablation studies to further study our choice of models and hyperparameters. Given its small size, we used SummHay for our ablation experiments.

\textbf{Gemini-1.5:} We run some preliminary experiments with Gemini-1.5 Flash and Pro (\autoref{tab:summhay_gemini}). Across methods, we consistently found that Gemini-1.5 models generate short summaries and underperform open source models. It is possible that we could improve their summaries using a different prompt, but we leave this extension to future work. Due to the high costs associated with Gemini API, we did not run experiments with our larger Background and WCEP datasets.

\textbf{Llama-3:} Our iterative methods do not require a long-context transformer, so we experiment with short-context transformers to see if they are better suited for this task. We run inference with Llama-3 8B and 70B (8k context window) in the SummHay and SummHay oracle settings (\autoref{tab:summhay_llama3}). We found that both models are either comparable or underperform their Llama-3.1 counterparts. It is likely that the Llama-3.1 models are better at short-text summarization.

\textbf{Chunk size:} As we have highlighted earlier, the hierarchical method exhibits a significant degradation in summary recall. We experiment with larger chunk sizes that allow for packing more intermediate summaries into the transformer. Our results using 8k, 16k and 32k chunk sizes show minimal improvements over our default 4k chunk size.

\textbf{Retriever:} Following the setup of SummHay \cite{laban-etal-2024-summhay}, we experiment with the E5-RoPE embedding for retrieval.\footnote{\url{https://huggingface.co/dwzhu/e5rope-base}} We report results in \autoref{tab:summhay_e5}. E5-RoPE performs slightly worse than the SFR-Embedding-2 results from \autoref{tab:summhay}.

\textbf{Grounded generation:} Jamba provides a grounded generation option in which the documents are passed as a separate object in the chat template. We experiment with this chat template to see if it provides any gains over our default setting of concatenating documents in the message. We report results in \autoref{tab:summhay_jamba_grounded}. Interestingly, this template helps improve the performance of hierarchical and incremental methods and hurts performance in full-context and retrieval settings. This needs further investigation. Command-R also includes a grounded generation template, but it is recommended for documents (or chunks) that contain 100-400 words. We couldn't make it work with full documents from our datasets.

\textbf{Filtered Background:} Our results showed that Background is the most challenging of the three datasets. To simplify the task, we pre-filter the documents using the update summary from the event timeline. We use the E5RoPE model \cite{zhu-etal-2024-longembed} to prefilter up to 128k tokens in the input for each example. However, we did not observe any significant improvements with this filtered dataset.

\input{tables/scores_summhay}
\input{tables/scores_summhayoracle}
\input{tables/scores_background}
\input{tables/scores_wcep}
\input{tables/summary_len_stats}

\input{tables/ablation_gemini}
\input{tables/ablation_llama3}
\input{tables/ablation_chunk_size}
\input{tables/ablation_retriever}
\input{tables/ablation_jamba}

\input{tables/qual_hierarchical}
\input{tables/qual_incremental}

\end{document}

%% file: tables/datasets.tex
\begin{table}[t]
    \centering
    \resizebox{0.45\textwidth}{!}{
    \begin{tabular}{@{}lrrrr@{}}
    \toprule
    & & & \multicolumn{2}{c}{Avg. length} \\
    Dataset & \# Ex. & \# Docs/Ex. & Doc. & Summ. \\
    \midrule
    SummHay & 92 & 100 & 884 & 185 \\
    Background & 658 & 186 & 1033 & 174 \\
    WCEP & 1020 & 76 & 468 & 34 \\
    \bottomrule
    \end{tabular}}
    \caption{An overview of our multi-document summarization datasets. We report the number of examples in the test set, and average statistics for \# documents per example, document and summary lengths (words).}
    \label{tab:datasets}
\end{table}

%% file: tables/scores_all.tex
\begin{table*}
\centering
\resizebox{0.8\textwidth}{!}{
\begin{tabular}{@{}m{2cm}@{ }m{3cm}@{ }m{3cm}@{ }m{3cm}@{ }m{3cm}@{ }m{3cm}@{}}
& Llama-3.1-8B & Llama-3.1-70B & Command-R & Jamba-1.5-Mini \\
\midrule
SummHay &
\centering
\begin{tikzpicture}[scale=0.5]
\draw (0,2) -- (4,2) node[right, font=\footnotesize] {33.9};
\draw[fill=blue] (0.5,2) rectangle (1,1.06);
\node[below, font=\scriptsize] at (0.75,1.06) {-53\%};
\draw[fill=orange] (2,2) rectangle (2.5,1.37);
\node[below, font=\scriptsize] at (2.25,1.37) {-32\%};
\draw[fill=magenta] (3.5,2) rectangle (4,2.04);
\node[above, font=\scriptsize] at (3.75,2.04) {+4\%};
\end{tikzpicture} &
\centering
\begin{tikzpicture}[scale=0.5]
\draw (0,2) -- (4,2) node[right, font=\footnotesize] {31.1};
\draw[fill=blue] (0.5,2) rectangle (1,1.11);
\node[below, font=\scriptsize] at (0.75,1.11) {-44\%};
\draw[fill=orange] (2,2) rectangle (2.5,1.27);
\node[below, font=\scriptsize] at (2.25,1.27) {-37\%};
\draw[fill=magenta] (3.5,2) rectangle (4,2.33);
\node[above, font=\scriptsize] at (3.75,2.33) {+17\%};
\end{tikzpicture} &
\centering
\begin{tikzpicture}[scale=0.5]
\draw (0,2) -- (4,2) node[right, font=\footnotesize] {30.4};
\draw[fill=blue] (0.5,2) rectangle (1,0.75);
\node[below, font=\scriptsize] at (0.75,0.75) {-63\%};
\draw[fill=orange] (2,2) rectangle (2.5,0.75);
\node[below, font=\scriptsize] at (2.25,0.75) {-63\%};
\draw[fill=magenta] (3.5,2) rectangle (4,2);
\node[above, font=\scriptsize] at (3.75,2) {0\%};
\end{tikzpicture} &
\begin{tikzpicture}[scale=0.5]
\draw (0,2) -- (4,2) node[right, font=\footnotesize] {32.9};
\draw[fill=blue] (0.5,2) rectangle (1,0.92);
\node[below, font=\scriptsize] at (0.75,0.92) {-54\%};
\draw[fill=orange] (2,2) rectangle (2.5,1.37);
\node[below, font=\scriptsize] at (2.25,1.37) {-32\%};
\draw[fill=magenta] (3.5,2) rectangle (4,2.11);
\node[above, font=\scriptsize] at (3.75,2.11) {+6\%};
\end{tikzpicture} \\
SummHay (oracle) &
\centering
\begin{tikzpicture}[scale=0.5]
\draw (0,2) -- (4,2) node[right, font=\footnotesize] {37.1};
\draw[fill=blue] (0.5,2) rectangle (1,1.45);
\node[below, font=\scriptsize] at (0.75,1.45) {-27\%};
\draw[fill=orange] (2,2) rectangle (2.5,1.67);
\node[below, font=\scriptsize] at (2.25,1.67) {-17\%};
\draw[fill=magenta] (3.5,2) rectangle (4,2.02);
\node[above, font=\scriptsize] at (3.75,2.02) {+1\%};
\end{tikzpicture} &
\centering
\begin{tikzpicture}[scale=0.5]
\draw (0,2) -- (4,2) node[right, font=\footnotesize] {41.8};
\draw[fill=blue] (0.5,2) rectangle (1,1.30);
\node[below, font=\scriptsize] at (0.75,1.30) {-35\%};
\draw[fill=orange] (2,2) rectangle (2.5,1.41);
\node[below, font=\scriptsize] at (2.25,1.41) {-29\%};
\draw[fill=magenta] (3.5,2) rectangle (4,1.88);
\node[below, font=\scriptsize] at (3.75,1.88) {-6\%};
\end{tikzpicture} &
\centering
\begin{tikzpicture}[scale=0.5]
\draw (0,2) -- (4,2) node[right, font=\footnotesize] {32.6};
\draw[fill=blue] (0.5,2) rectangle (1,1.33);
\node[below, font=\scriptsize] at (0.75,1.33) {-33\%};
\draw[fill=orange] (2,2) rectangle (2.5,1.05);
\node[below, font=\scriptsize] at (2.25,1.05) {-47\%};
\draw[fill=magenta] (3.5,2) rectangle (4,1.95);
\node[below, font=\scriptsize] at (3.75,1.95) {-3\%};
\end{tikzpicture} &
\begin{tikzpicture}[scale=0.5]
\draw (0,2) -- (4,2) node[right, font=\footnotesize] {35.1};
\draw[fill=blue] (0.5,2) rectangle (1,1.69);
\node[below, font=\scriptsize] at (0.75,1.69) {-16\%};
\draw[fill=orange] (2,2) rectangle (2.5,1.65);
\node[below, font=\scriptsize] at (2.25,1.65) {-18\%};
\draw[fill=magenta] (3.5,2) rectangle (4,2.07);
\node[above, font=\scriptsize] at (3.75,2.07) {+3\%};
\end{tikzpicture} \\
Background &
\centering
\begin{tikzpicture}[scale=0.5]
\draw (0,2) -- (4,2) node[right, font=\footnotesize] {15.6};
\draw[fill=blue] (0.5,2) rectangle (1,1.68);
\node[below, font=\scriptsize] at (0.75,1.68) {-16\%};
\draw[fill=orange] (2,2) rectangle (2.5,1.28);
\node[below, font=\scriptsize] at (2.25,1.28) {-36\%};
\draw[fill=magenta] (3.5,2) rectangle (4,2.17);
\node[above, font=\scriptsize] at (3.75,2.17) {+8\%};
\end{tikzpicture} &
\centering
\begin{tikzpicture}[scale=0.5]
\draw (0,2) -- (4,2) node[right, font=\footnotesize] {16.1};
\draw[fill=blue] (0.5,2) rectangle (1,1.83);
\node[below, font=\scriptsize] at (0.75,1.83) {-9\%};
\draw[fill=orange] (2,2) rectangle (2.5,1.38);
\node[below, font=\scriptsize] at (2.25,1.38) {-31\%};
\draw[fill=magenta] (3.5,2) rectangle (4,1.95);
\node[below, font=\scriptsize] at (3.75,1.95) {-3\%};
\end{tikzpicture} &
\centering
\begin{tikzpicture}[scale=0.5]
\draw (0,2) -- (4,2) node[right, font=\footnotesize] {10.3};
\draw[fill=blue] (0.5,2) rectangle (1,1.81);
\node[below, font=\scriptsize] at (0.75,1.81) {-10\%};
\draw[fill=orange] (2,2) rectangle (2.5,1.88);
\node[below, font=\scriptsize] at (2.25,1.88) {-6\%};
\draw[fill=magenta] (3.5,2) rectangle (4,2.29);
\node[above, font=\scriptsize] at (3.75,2.29) {+15\%};
\end{tikzpicture} &
\begin{tikzpicture}[scale=0.5]
\draw (0,2) -- (4,2) node[right, font=\footnotesize] {12.3};
\draw[fill=blue] (0.5,2) rectangle (1,1.66);
\node[below, font=\scriptsize] at (0.75,1.66) {-17\%};
\draw[fill=orange] (2,2) rectangle (2.5,1.69);
\node[below, font=\scriptsize] at (2.25,1.69) {-15\%};
\draw[fill=magenta] (3.5,2) rectangle (4,1.79);
\node[below, font=\scriptsize] at (3.75,1.79) {-11\%};
\end{tikzpicture} \\
WCEP &
\centering
\begin{tikzpicture}[scale=0.5]
\draw (0,2) -- (4,2) node[right, font=\footnotesize] {30.7};
\draw[fill=blue] (0.5,2) rectangle (1,1.71);
\node[below, font=\scriptsize] at (0.75,1.71) {-15\%};
\draw[fill=orange] (2,2) rectangle (2.5,1.56);
\node[below, font=\scriptsize] at (2.25,1.56) {-22\%};
\draw[fill=magenta] (3.5,2) rectangle (4,1.94);
\node[below, font=\scriptsize] at (3.75,1.94) {-3\%};
\end{tikzpicture} &
\centering
\begin{tikzpicture}[scale=0.5]
\draw (0,2) -- (4,2) node[right, font=\footnotesize] {31.1};
\draw[fill=blue] (0.5,2) rectangle (1,1.74);
\node[below, font=\scriptsize] at (0.75,1.74) {-13\%};
\draw[fill=orange] (2,2) rectangle (2.5,1.56);
\node[below, font=\scriptsize] at (2.25,1.56) {-22\%};
\draw[fill=magenta] (3.5,2) rectangle (4,1.96);
\node[below, font=\scriptsize] at (3.75,1.96) {-2\%};
\end{tikzpicture} &
\centering
\begin{tikzpicture}[scale=0.5]
\draw (0,2) -- (4,2) node[right, font=\footnotesize] {28.9};
\draw[fill=blue] (0.5,2) rectangle (1,1.77);
\node[below, font=\scriptsize] at (0.75,1.77) {-11\%};
\draw[fill=orange] (2,2) rectangle (2.5,1.74);
\node[below, font=\scriptsize] at (2.25,1.74) {-13\%};
\draw[fill=magenta] (3.5,2) rectangle (4,1.97);
\node[below, font=\scriptsize] at (3.75,1.97) {-1\%};
\end{tikzpicture} &
\begin{tikzpicture}[scale=0.5]
\draw (0,2) -- (4,2) node[right, font=\footnotesize] {29.3};
\draw[fill=blue] (0.5,2) rectangle (1,1.96);
\node[below, font=\scriptsize] at (0.75,1.96) {-2\%};
\draw[fill=orange] (2,2) rectangle (2.5,1.73);
\node[below, font=\scriptsize] at (2.25,1.73) {-13\%};
\draw[fill=magenta] (3.5,2) rectangle (4,1.99);
\node[below, font=\scriptsize] at (3.75,1.99) {-1\%};
\end{tikzpicture} \\
\bottomrule
\end{tabular}}
\caption{Performance of {\color{blue} hierarchical}, {\color{orange}incremental} and {\color{magenta}retrieval} methods relative to the full-context baseline.}
\label{tab:overall_a3cu}
\end{table*}

%% file: figures/inf_loss.tex
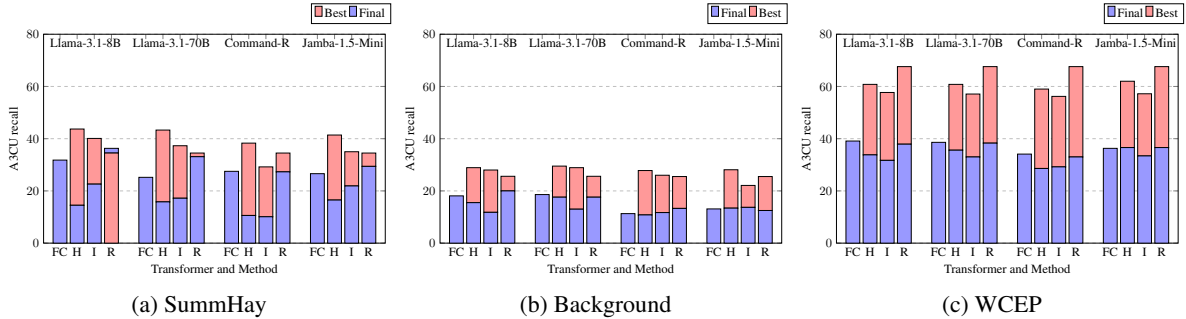
\begin{figure*}[t]
    \centering
    \begin{subfigure}[b]{0.32\textwidth}
    \resizebox{\textwidth}{!}{\input{figures/inf_loss_summhay}}
    \caption{SummHay}
    \label{fig:inf_loss_summhay}
    \end{subfigure}
    \begin{subfigure}[b]{0.32\textwidth}
    \resizebox{\textwidth}{!}{\input{figures/inf_loss_background}}
    \caption{Background}
    \label{fig:inf_loss_background}
    \end{subfigure}
    \begin{subfigure}[b]{0.32\textwidth}
    \resizebox{\textwidth}{!}{\input{figures/inf_loss_wcep}}
    \caption{WCEP}
    \label{fig:inf_loss_wcep}
    \end{subfigure}
    \caption{Salient information retention in the intermediate and final summaries (A3CU \emph{recall}). For each compression method, we report the best recall from the intermediate outputs and the recall of the final summary. (H: hierarchical, I: incremental, R: retrieval, FC: full-context)}
    \label{fig:inf_loss}
\end{figure*}

%% file: figures/inf_loss_summhay.tex
\begin{tikzpicture}
    \begin{axis}[
     width=12cm,
     height=8cm,
     ybar stacked,
     bar width=12pt,
     ylabel={A3CU recall},
     xlabel={Transformer and Method},
     xticklabels={
     FC, H, I, R,
     FC, H, I, R,
     FC, H, I, R,
     FC, H, I, R
     },
     xtick={0,1,2,3,5,6,7,8,10,11,12,13,15,16,17,18},
     x tick label style={anchor=north},
     legend style={at={(0.9,1.05)},anchor=south,legend columns=-1},
     ymajorgrids=true,
     grid style=dashed,
     ymin=0,
     ymax=80,
     enlarge x limits={abs=0.5cm},
    ]
    \addplot[fill=red!40] coordinates {
     (0,0) (1,0) (2,0) (3,34.5)
     (5,0) (6,0) (7,0) (8,0)
     (10,0) (11,0) (12,0) (13,0)
     (15,0) (16,0) (17,0) (18,0)
     };
    \addplot[fill=blue!40] coordinates {
     (0,31.8) (1,14.5) (2,22.6) (3,1.8)
     (5,25.2) (6,15.8) (7,17.2) (8,33.1)
     (10,27.5) (11,10.6) (12,10.1) (13,27.3)
     (15,26.6) (16,16.5) (17,21.9) (18,29.4)
     };
    \addplot[fill=red!40] coordinates {
     (0,0) (1,29.2) (2,17.5) (3,0)
     (5,0) (6,27.5) (7,20.1) (8,1.4)
     (10,0) (11,27.7) (12,19.1) (13,7.2)
     (15,0) (16,24.9) (17,13.1) (18,5.1)
     };
    \legend{Best,Final}
    \node[anchor=north] at (axis cs:1.5,80) {Llama-3.1-8B};
    \node[anchor=north] at (axis cs:6.5,80) {Llama-3.1-70B};
    \node[anchor=north] at (axis cs:11.5,80) {Command-R};
    \node[anchor=north] at (axis cs:16.5,80) {Jamba-1.5-Mini};
    \end{axis}
\end{tikzpicture}

%% file: figures/inf_loss_background.tex
\begin{tikzpicture}
    \begin{axis}[
     width=12cm,
     height=8cm,
     ybar stacked,
     bar width=12pt,
     ylabel={A3CU recall},
     xlabel={Transformer and Method},
     xticklabels={
     FC, H, I, R,
     FC, H, I, R,
     FC, H, I, R,
     FC, H, I, R
     },
     xtick={0,1,2,3,5,6,7,8,10,11,12,13,15,16,17,18},
     x tick label style={anchor=north},
     legend style={at={(0.9,1.05)},anchor=south,legend columns=-1},
     ymajorgrids=true,
     grid style=dashed,
     ymin=0,
     ymax=80,
     enlarge x limits={abs=0.5cm},
    ]
    \addplot[fill=blue!40] coordinates {
     (0,18.1) (1,15.5) (2,11.8) (3,20.0)
     (5,18.6) (6,17.6) (7,13.0) (8,17.6)
     (10,11.3) (11,10.8) (12,11.7) (13,13.3)
     (15,13.1) (16,13.4) (17,13.7) (18,12.5)
     };
    \addplot[fill=red!40] coordinates {
     (0,0) (1,13.4) (2,16.2) (3,5.6)
     (5,0) (6,11.9) (7,15.9) (8,8.0)
     (10,0) (11,17.0) (12,14.3) (13,12.2)
     (15,0) (16,14.7) (17,8.4) (18,13.0)
     };
    \legend{Final, Best}
    \node[anchor=north] at (axis cs:1.5,80) {Llama-3.1-8B};
    \node[anchor=north] at (axis cs:6.5,80) {Llama-3.1-70B};
    \node[anchor=north] at (axis cs:11.5,80) {Command-R};
    \node[anchor=north] at (axis cs:16.5,80) {Jamba-1.5-Mini};
    \end{axis}
\end{tikzpicture}

%% file: figures/inf_loss_wcep.tex
\begin{tikzpicture}
    \begin{axis}[
     width=12cm,
     height=8cm,
     ybar stacked,
     bar width=12pt,
     ylabel={A3CU recall},
     xlabel={Transformer and Method},
     xticklabels={
     FC, H, I, R,
     FC, H, I, R,
     FC, H, I, R,
     FC, H, I, R
     },
     xtick={0,1,2,3,5,6,7,8,10,11,12,13,15,16,17,18},
     x tick label style={anchor=north},
     legend style={at={(0.9,1.05)},anchor=south,legend columns=-1},
     ymajorgrids=true,
     grid style=dashed,
     ymin=0,
     ymax=80,
     enlarge x limits={abs=0.5cm},
    ]
    \addplot[fill=blue!40] coordinates {
     (0,39.1) (1,33.8) (2,31.7) (3,37.9)
     (5,38.6) (6,35.6) (7,33.0) (8,38.3)
     (10,34.1) (11,28.6) (12,29.2) (13,33.0)
     (15,36.3) (16,36.6) (17,33.4) (18,36.6)
     };
    \addplot[fill=red!40] coordinates {
     (0,0) (1,27.0) (2,26.0) (3,29.7)
     (5,0) (6,25.2) (7,24.1) (8,29.3)
     (10,0) (11,30.4) (12,27.0) (13,34.6)
     (15,0) (16,25.4) (17,23.8) (18,31.0)
     };
    \legend{Final, Best}
    \node[anchor=north] at (axis cs:1.5,80) {Llama-3.1-8B};
    \node[anchor=north] at (axis cs:6.5,80) {Llama-3.1-70B};
    \node[anchor=north] at (axis cs:11.5,80) {Command-R};
    \node[anchor=north] at (axis cs:16.5,80) {Jamba-1.5-Mini};
    \end{axis}
\end{tikzpicture}

%% file: tables/human_eval.tex
\begin{table}[t]
    \centering
    \begin{tabular}{@{}llrr@{}}
    \toprule
    Transformer & Method & Best & Worst \\
    \midrule
    Llama-3.1-8B & Full-Context & 28 & 10 \\
    Llama-3.1-8B & Hierarchical & 13 & 44 \\
    Llama-3.1-8B & Incremental & 18 & 21 \\
    Llama-3.1-8B & Retrieval & 45 & 4 \\
    \bottomrule
    \end{tabular}
    \caption{Best-worst ratings from human evaluation on a random sample of 62 examples from SummHay. We report the counts for number of times a system was rated the best or worst amongst the four summaries. We compare each system summary against the reference.}
    \label{tab:human_eval}
\end{table}

%% file: tables/prompts.tex
\begin{table}
    \centering
    \resizebox{0.48\textwidth}{!}{
    \begin{tabular}{p{9cm}}
    \toprule
    \texttt{\{document\}} \\
    \\
    Question: \texttt{\{question\}} \\
    \\
    Answer the question based on the provided document. Be concise and directly address only the specific question asked. Limit your response to a maximum of \texttt{\{num\_words\}} words. \\
    \bottomrule
    \end{tabular}}
    \caption{Prompt for our summarization task. We pass the input documents concatenated together by a \textbackslash n character. The number of words in the summary are determined by the dataset (\autoref{tab:datasets}).}
    \label{tab:summarization_prompt}
\end{table}

%% file: figures/inf_loss_appendix.tex
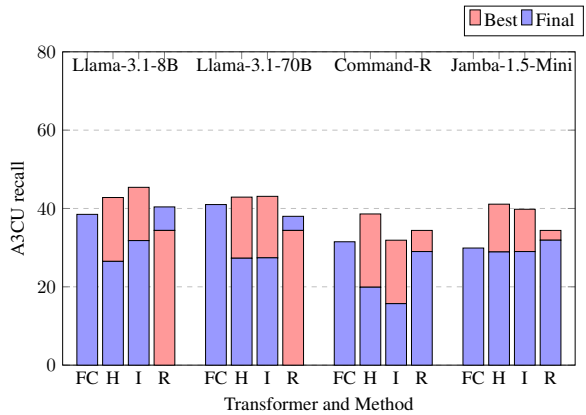
\begin{figure}[t]
    \centering
    \resizebox{0.48\textwidth}{!}{\input{figures/inf_loss_summhayoracle}}
    \caption{Salient information retention in the intermediate and final summaries (A3CU \emph{recall}) for SummHay (oracle). For each compression method, we report the best recall from the intermediate outputs and the recall of the final summary. (H: hierarchical, I: incremental, R: retrieval, FC: full-context)}
    \label{fig:inf_loss_summhayoracle}
\end{figure}

%% file: figures/inf_loss_summhayoracle.tex
\begin{tikzpicture}
    \begin{axis}[
        width=12cm,
        height=8cm,
        ybar stacked,
        bar width=12pt,
        ylabel={A3CU recall},
        xlabel={Transformer and Method},
        xticklabels={
            FC, H, I, R,
            FC, H, I, R,
            FC, H, I, R,
            FC, H, I, R
        },
        xtick={0,1,2,3,5,6,7,8,10,11,12,13,15,16,17,18},
        x tick label style={anchor=north},
        legend style={at={(0.9,1.05)},anchor=south,legend columns=-1},
        ymajorgrids=true,
        grid style=dashed,
        ymin=0,
        ymax=80,
        enlarge x limits={abs=0.5cm},
    ]
    \addplot[fill=red!40] coordinates {
         (0,0) (1,0) (2,0) (3,34.4)
         (5,0) (6,0) (7,0) (8,34.4)
         (10,0) (11,0) (12,0) (13,0)
         (15,0) (16,0) (17,0) (18,0)
     };
    \addplot[fill=blue!40] coordinates {
        (0,38.5) (1,26.5) (2,31.8) (3,6.0)
        (5,41.0) (6,27.3) (7,27.4) (8,3.6)
        (10,31.5) (11,19.9) (12,15.7) (13,29.0)
        (15,29.9) (16,28.9) (17,29.0) (18,31.9)
    };
    \addplot[fill=red!40] coordinates {
        (0,0) (1,16.3) (2,13.6) (3,0)
        (5,0) (6,15.6) (7,15.7) (8,0)
        (10,0) (11,18.7) (12,16.2) (13,5.4)
        (15,0) (16,12.2) (17,10.8) (18,2.5)
    };
    \legend{Best,Final}
    \node[anchor=north] at (axis cs:1.5,80) {Llama-3.1-8B};
    \node[anchor=north] at (axis cs:6.5,80) {Llama-3.1-70B};
    \node[anchor=north] at (axis cs:11.5,80) {Command-R};
    \node[anchor=north] at (axis cs:16.5,80) {Jamba-1.5-Mini};

    \end{axis}
\end{tikzpicture}

%% file: figures/box_a3cu.tex
\begin{figure*}[t]
    \centering
    \begin{subfigure}{0.49\textwidth}
    \includegraphics[width=\textwidth]{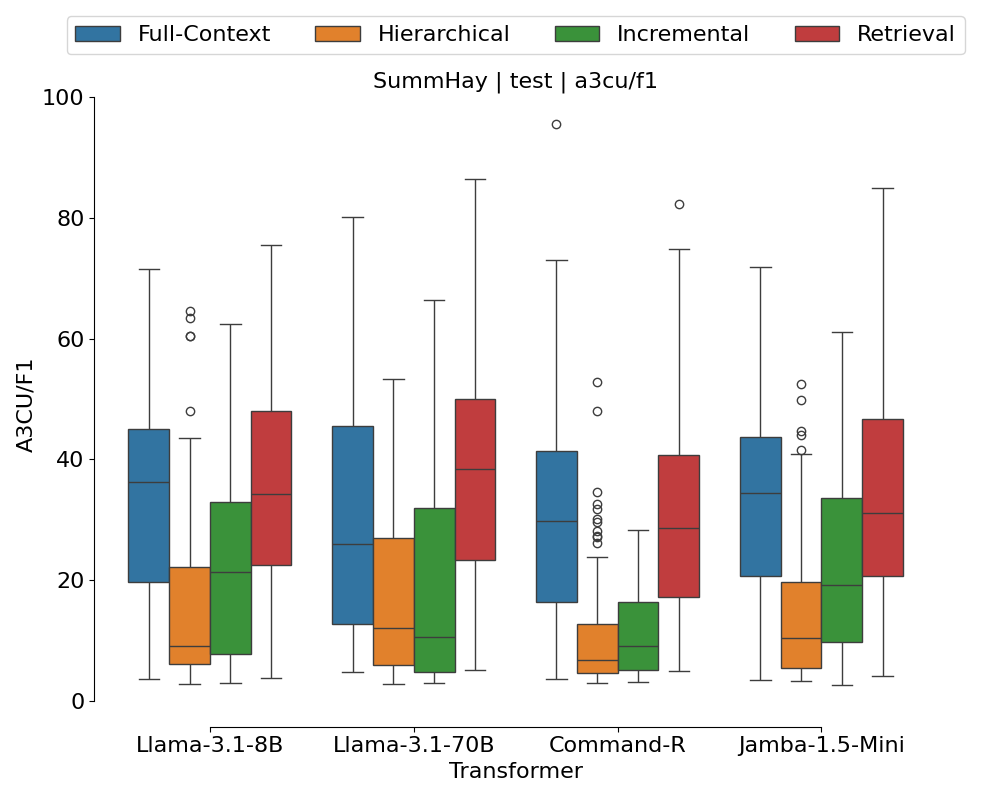}
    \end{subfigure}
    \begin{subfigure}{0.49\textwidth}
    \includegraphics[width=\textwidth]{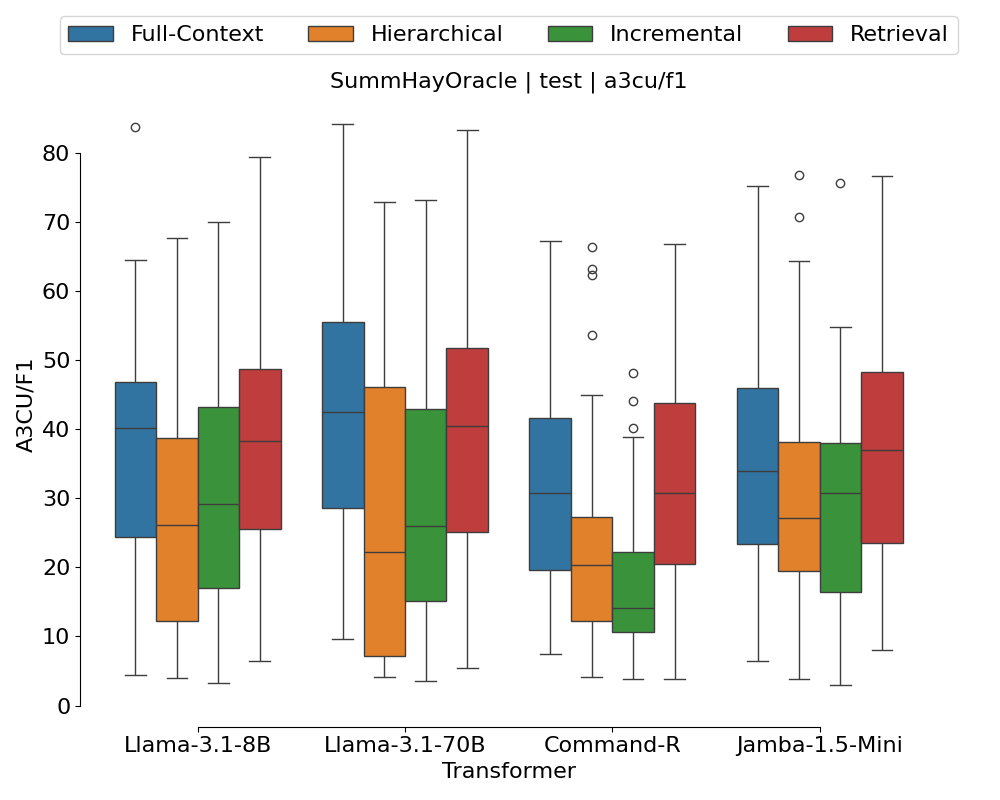}    
    \end{subfigure}
    \begin{subfigure}{0.49\textwidth}
    \includegraphics[width=\textwidth]{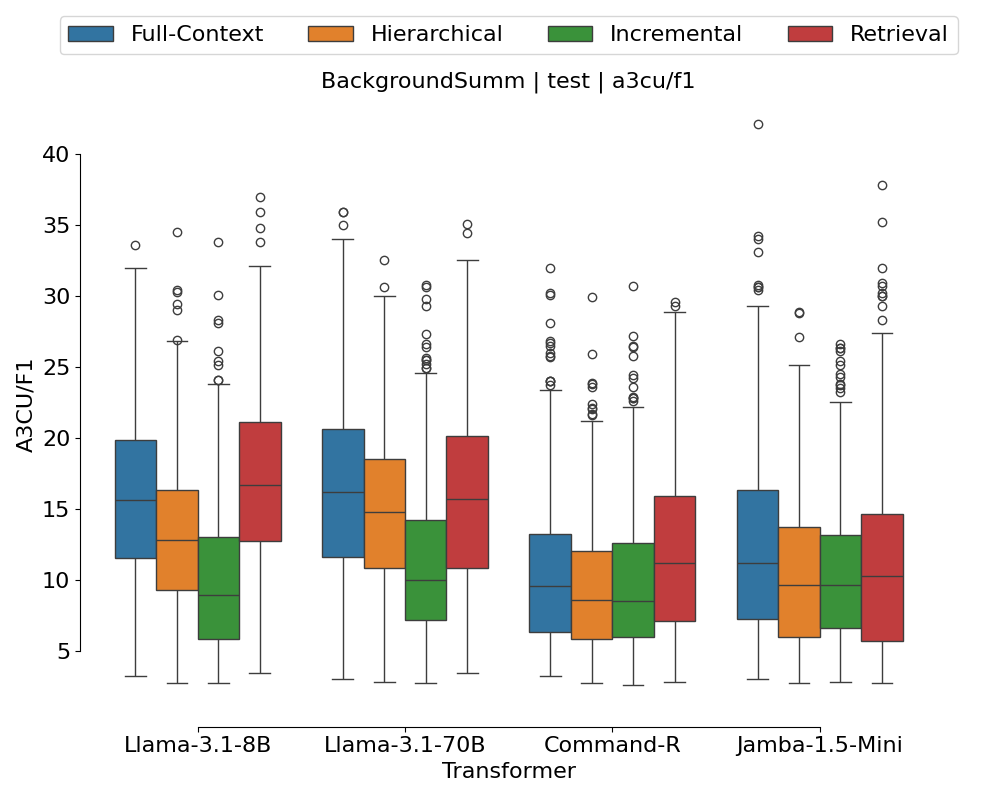}
    \end{subfigure}
    \begin{subfigure}{0.49\textwidth}
    \includegraphics[width=\textwidth]{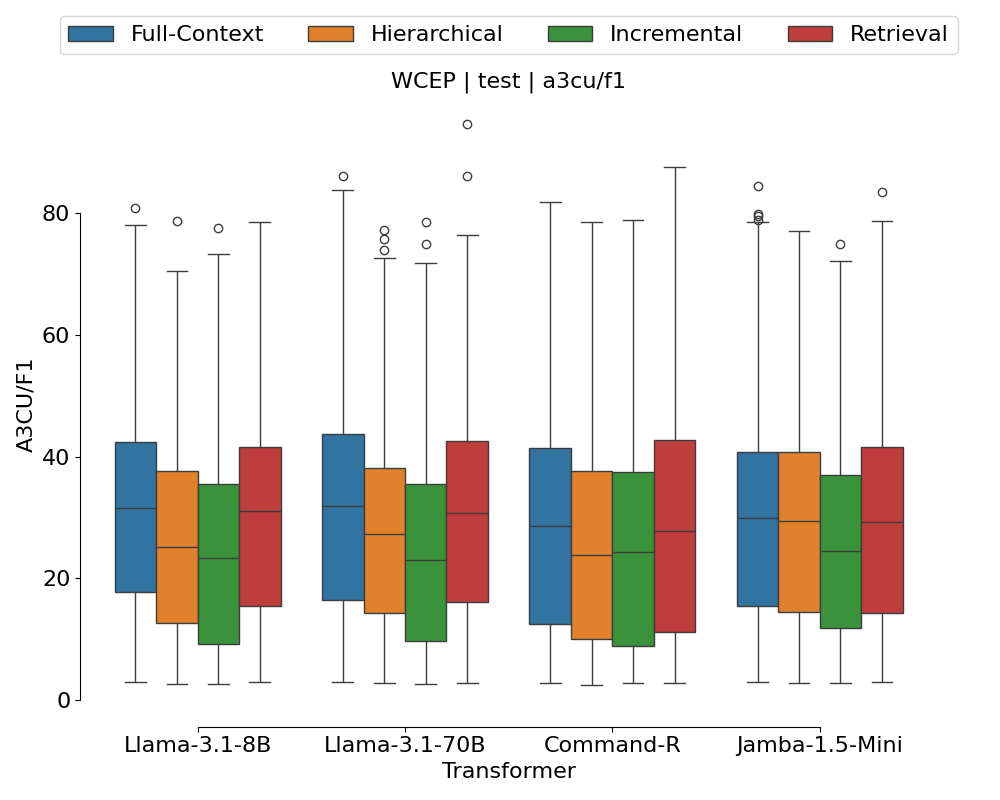}
    \end{subfigure}
    \caption{A3CU F1 score distribution across examples.}
    \label{fig:box_a3cu}
\end{figure*}

%% file: tables/scores_summhay.tex
\begin{table*}
\centering
\resizebox{\textwidth}{!}{
\begin{tabular}{@{}llrrrrrrr@{}}
\toprule
& & & & & & \multicolumn{3}{c}{A3CU} \\
Transformer & Method & ROUGE-1 & ROUGE-2 & ROUGE-L & ROUGE-Lsum & Recall & Precision & F1 \\
\midrule
Llama-3.1-8B & Full-Context & 49.4 & 25.4 & 28.5 & 46.4 & 31.8 & 39.5 & 33.9 \\
Llama-3.1-8B & Hierarchical & 29.4 & 10.8 & 16.4 & 27.1 & 14.5 & 23.3 & 16.0 \\
Llama-3.1-8B & Incremental & 41.5 & 16.4 & 22.5 & 38.0 & 22.6 & 27.5 & 23.2 \\
Llama-3.1-8B & Retrieval & 51.8 & 27.0 & 29.3 & 48.9 & 36.3 & 36.7 & 35.3 \\
\midrule
Llama-3.1-70B & Full-Context & 43.7 & 23.8 & 25.9 & 41.3 & 25.2 & 46.3 & 31.1 \\
Llama-3.1-70B & Hierarchical & 30.0 & 11.0 & 16.4 & 27.2 & 15.8 & 23.6 & 17.3 \\
Llama-3.1-70B & Incremental & 33.1 & 13.6 & 19.3 & 30.5 & 17.2 & 27.5 & 19.7 \\
Llama-3.1-70B & Retrieval & 50.2 & 26.7 & 29.3 & 47.1 & 33.1 & 43.8 & 36.3 \\
\midrule
Command-R & Full-Context & 45.0 & 19.0 & 24.4 & 41.2 & 27.5 & 38.1 & 30.4 \\
Command-R & Hierarchical & 35.4 & 8.0 & 18.4 & 32.0 & 10.6 & 13.9 & 11.4 \\
Command-R & Incremental & 33.0 & 7.7 & 17.8 & 29.7 & 10.1 & 15.9 & 11.4 \\
Command-R & Retrieval & 45.0 & 19.6 & 24.9 & 41.8 & 27.3 & 38.3 & 30.4 \\
\midrule
Jamba-1.5-Mini & Full-Context & 44.2 & 22.0 & 27.0 & 41.2 & 26.6 & 47.7 & 32.9 \\
Jamba-1.5-Mini & Hierarchical & 38.1 & 11.6 & 19.2 & 35.0 & 16.5 & 15.9 & 15.1 \\
Jamba-1.5-Mini & Incremental & 40.7 & 15.9 & 21.8 & 37.1 & 21.9 & 27.8 & 22.5 \\
Jamba-1.5-Mini & Retrieval & 46.4 & 22.8 & 27.6 & 42.8 & 29.4 & 46.4 & 34.7 \\
\bottomrule
\end{tabular}}
\caption{Results on SummHay.}
\label{tab:summhay}
\end{table*}

%% file: tables/scores_summhayoracle.tex
\begin{table*}[t]
\centering
\resizebox{\textwidth}{!}{
\begin{tabular}{@{}llrrrrrrr@{}}
\toprule
& & & & & & \multicolumn{3}{c}{A3CU} \\
Transformer & Method & ROUGE-1 & ROUGE-2 & ROUGE-L & ROUGE-Lsum & Recall & Precision & F1 \\
\midrule
Llama-3.1-8B & Full-Context & 53.4 & 29.0 & 29.7 & 50.1 & 38.5 & 37.9 & 37.1 \\
Llama-3.1-8B & Hierarchical & 40.7 & 18.2 & 21.4 & 38.0 & 26.5 & 31.9 & 27.0 \\
Llama-3.1-8B & Incremental & 48.0 & 21.8 & 25.2 & 44.6 & 31.8 & 32.9 & 30.9 \\
Llama-3.1-8B & Retrieval & 53.7 & 28.8 & 29.8 & 50.5 & 40.4 & 37.2 & 37.5 \\
\midrule
Llama-3.1-70B & Full-Context & 54.1 & 30.1 & 30.7 & 51.0 & 41.0 & 45.8 & 41.8 \\
Llama-3.1-70B & Hierarchical & 37.6 & 18.3 & 21.1 & 34.9 & 27.3 & 32.3 & 27.2 \\
Llama-3.1-70B & Incremental & 41.8 & 20.2 & 23.5 & 38.7 & 27.4 & 37.8 & 29.5 \\
Llama-3.1-70B & Retrieval & 53.3 & 28.7 & 30.1 & 50.3 & 38.0 & 44.0 & 39.3 \\
\midrule
Command-R & Full-Context & 48.3 & 20.2 & 25.4 & 44.2 & 31.5 & 38.0 & 32.6 \\
Command-R & Hierarchical & 41.7 & 12.5 & 21.3 & 38.1 & 19.9 & 26.8 & 21.7 \\
Command-R & Incremental & 37.1 & 11.0 & 19.8 & 33.3 & 15.7 & 22.6 & 17.2 \\
Command-R & Retrieval & 46.5 & 19.9 & 25.1 & 42.7 & 29.0 & 38.6 & 31.8 \\
\midrule
Jamba-1.5-Mini & Full-Context & 47.6 & 24.3 & 28.2 & 44.4 & 29.9 & 47.8 & 35.1 \\
Jamba-1.5-Mini & Hierarchical & 46.7 & 20.3 & 25.6 & 43.5 & 28.9 & 33.5 & 29.6 \\
Jamba-1.5-Mini & Incremental & 46.2 & 20.5 & 24.4 & 42.9 & 29.0 & 32.5 & 28.9 \\
Jamba-1.5-Mini & Retrieval & 48.5 & 24.7 & 28.0 & 45.2 & 31.9 & 46.2 & 36.3 \\
\bottomrule
\end{tabular}}
\caption{Results on SummHay (oracle).}
\label{tab:summhay_oracle}
\end{table*}

%% file: tables/scores_background.tex
\begin{table*}[t]
\centering
\resizebox{\textwidth}{!}{
\begin{tabular}{@{}llrrrrrrr@{}}
\toprule
& & & & & & \multicolumn{3}{c}{A3CU} \\
Transformer & Method & ROUGE-1 & ROUGE-2 & ROUGE-L & ROUGE-Lsum & Recall & Precision & F1 \\
\midrule
Llama-3.1-8B & Full-Context & 36.5 & 8.4 & 18.3 & 33.2 & 18.1 & 15.4 & 15.6 \\
Llama-3.1-8B & Hierarchical & 35.2 & 7.2 & 17.5 & 32.0 & 15.5 & 12.8 & 13.1 \\
Llama-3.1-8B & Incremental & 34.4 & 6.6 & 16.4 & 31.1 & 11.8 & 10.5 & 10.0 \\
Llama-3.1-8B & Retrieval & 37.7 & 8.7 & 19.0 & 34.2 & 20.0 & 16.2 & 16.9 \\
\midrule
Llama-3.1-70B & Full-Context & 36.6 & 8.7 & 18.4 & 33.4 & 18.6 & 15.8 & 16.1 \\
Llama-3.1-70B & Hierarchical & 34.5 & 7.5 & 17.4 & 31.4 & 17.6 & 14.2 & 14.7 \\
Llama-3.1-70B & Incremental & 35.2 & 7.2 & 16.5 & 31.9 & 13.0 & 11.6 & 11.1 \\
Llama-3.1-70B & Retrieval & 35.7 & 8.0 & 18.6 & 32.2 & 17.6 & 16.0 & 15.7 \\
\midrule
Command-R & Full-Context & 31.9 & 6.1 & 17.5 & 28.6 & 11.3 & 11.4 & 10.3 \\
Command-R & Hierarchical & 31.5 & 5.8 & 16.7 & 28.7 & 10.8 & 9.5 & 9.3 \\
Command-R & Incremental & 34.6 & 6.7 & 16.3 & 31.3 & 11.7 & 9.9 & 9.7 \\
Command-R & Retrieval & 33.2 & 6.4 & 17.2 & 29.9 & 13.3 & 12.0 & 11.8 \\
\midrule
Jamba-1.5-Mini & Full-Context & 33.6 & 6.8 & 17.7 & 30.1 & 13.1 & 14.2 & 12.3 \\
Jamba-1.5-Mini & Hierarchical & 33.5 & 6.0 & 16.1 & 30.4 & 13.4 & 9.2 & 10.2 \\
Jamba-1.5-Mini & Incremental & 35.5 & 6.7 & 16.2 & 32.1 & 13.7 & 9.8 & 10.4 \\
Jamba-1.5-Mini & Retrieval & 33.0 & 6.1 & 16.8 & 29.5 & 12.5 & 11.8 & 11.0 \\
\bottomrule
\end{tabular}}
\caption{Results on Background.}
\label{tab:background}
\end{table*}

%% file: tables/scores_wcep.tex
\begin{table*}[t]
\centering
\resizebox{\textwidth}{!}{
\begin{tabular}{@{}llrrrrrrr@{}}
\toprule
& & & & & & \multicolumn{3}{c}{A3CU} \\
Transformer & Method & ROUGE-1 & ROUGE-2 & ROUGE-L & ROUGE-Lsum & Recall & Precision & F1 \\
\midrule
Llama-3.1-8B & Full-Context & 37.5 & 14.2 & 26.4 & 29.6 & 39.1 & 29.2 & 30.7 \\
Llama-3.1-8B & Hierarchical & 33.9 & 11.3 & 23.8 & 26.1 & 33.8 & 25.3 & 26.2 \\
Llama-3.1-8B & Incremental & 32.7 & 10.5 & 22.8 & 25.6 & 31.7 & 22.9 & 24.0 \\
Llama-3.1-8B & Retrieval & 36.8 & 13.7 & 26.1 & 29.0 & 37.9 & 28.4 & 29.7 \\
\midrule
Llama-3.1-70B & Full-Context & 37.5 & 14.1 & 26.7 & 30.0 & 38.6 & 30.7 & 31.1 \\
Llama-3.1-70B & Hierarchical & 34.3 & 11.4 & 23.8 & 26.6 & 35.6 & 25.7 & 27.1 \\
Llama-3.1-70B & Incremental & 32.5 & 10.4 & 22.6 & 25.5 & 33.0 & 22.7 & 24.2 \\
Llama-3.1-70B & Retrieval & 37.5 & 14.2 & 26.6 & 30.0 & 38.3 & 29.8 & 30.5 \\
\midrule
Command-R & Full-Context & 36.6 & 13.7 & 26.1 & 29.9 & 34.1 & 30.2 & 28.9 \\
Command-R & Hierarchical & 34.1 & 11.1 & 23.9 & 26.4 & 28.6 & 28.4 & 25.6 \\
Command-R & Incremental & 34.3 & 11.7 & 24.2 & 27.4 & 29.2 & 27.0 & 25.1 \\
Command-R & Retrieval & 36.7 & 13.7 & 26.0 & 29.7 & 33.0 & 29.8 & 28.5 \\
\midrule
Jamba-1.5-Mini & Full-Context & 36.8 & 13.8 & 25.8 & 29.8 & 36.3 & 28.6 & 29.3 \\
Jamba-1.5-Mini & Hierarchical & 35.8 & 12.8 & 25.1 & 28.8 & 36.6 & 27.9 & 28.7 \\
Jamba-1.5-Mini & Incremental & 34.3 & 11.7 & 23.6 & 27.7 & 33.4 & 24.2 & 25.4 \\
Jamba-1.5-Mini & Retrieval & 36.7 & 13.7 & 25.6 & 29.4 & 36.6 & 28.3 & 29.1 \\
\bottomrule
\end{tabular}}
\caption{Results on WCEP.}
\label{tab:wcep}
\end{table*}

%% file: tables/summary_len_stats.tex
\begin{table*}[t]
    \centering
    \begin{tabular}{lrrrrrr}
    \toprule
    & Full Context & Retrieval & \multicolumn{2}{r}{Hierarchical} & \multicolumn{2}{r}{Incremental}\\
    & & & Best & Final & Best & Final \\
    \midrule
    \multicolumn{7}{c}{SummHay (Reference: 185)} \\
    \midrule
    Llama-3.1-8B & 162 & 195 & 172 & 106 & 171 & 141 \\
    Llama-3.1-70B & 106 & 148 & 161 & 113 & 150 & 93 \\
    Command-R & 135 & 134 & 165 & 151 & 161 & 115 \\
    Jamba-1.5-Mini & 110 & 120 & 163 & 211 & 177 & 145 \\
    \midrule
    \multicolumn{7}{c}{Background (Reference: 174)} \\
    \midrule
    Llama-3.1-8B & 228 & 232 & 214 & 222 & 212 & 206 \\
    Llama-3.1-70B & 232 & 219 & 208 & 210 & 210 & 205 \\
    Command-R & 190 & 215 & 226 & 227 & 236 & 232 \\
    Jamba-1.5-Mini & 162 & 183 & 213 & 237 & 230 & 233 \\
    \midrule
    \multicolumn{7}{c}{WCEP (Reference: 35)} \\
    \midrule
    Llama-3.1-8B & 44 & 44 & 43 & 41 & 43 & 43 \\
    Llama-3.1-70B & 42 & 42 & 43 & 42 & 44 & 43 \\
    Command-R & 42 & 41 & 42 & 39 & 42 & 41 \\
    Jamba-1.5-Mini & 45 & 45 & 45 & 44 & 45 & 44 \\
    \bottomrule
    \end{tabular}
    \caption{Summary length statistics, using NLTK word tokenizer.}
    \label{tab:summary_len_stats}
\end{table*}

%% file: tables/ablation_gemini.tex
\begin{table*}
\centering
\resizebox{\textwidth}{!}{
\begin{tabular}{@{}llrrrrrrr@{}}
\toprule
& & & & & & \multicolumn{3}{c}{A3CU} \\
Transformer & Method & ROUGE-1 & ROUGE-2 & ROUGE-L & ROUGE-Lsum & Recall & Precision & F1 \\
\midrule
Gemini-1.5-Flash & Full-Context & 32.3 & 15.1 & 19.7 & 29.8 & 19.2 & 40.6 & 24.6 \\
Gemini-1.5-Flash & Hierarchical & 12.5 & 4.5 & 7.2 & 11.2 & 8.0 & 17.2 & 10.2 \\
Gemini-1.5-Flash & Incremental & 37.2 & 15.5 & 21.7 & 34.2 & 19.6 & 34.8 & 23.8 \\
Gemini-1.5-Flash & Retrieval & 37.5 & 18.7 & 23.3 & 34.8 & 22.4 & 47.4 & 28.3 \\
\midrule
Gemini-1.5-Pro & Full-Context & 41.8 & 18.3 & 23.9 & 38.8 & 26.2 & 36.8 & 29.2 \\
Gemini-1.5-Pro & Hierarchical & 10.9 & 3.1 & 6.5 & 9.7 & 6.9 & 17.0 & 9.2 \\
Gemini-1.5-Pro & Incremental & 22.7 & 6.4 & 13.4 & 20.4 & 10.3 & 21.8 & 12.9 \\
Gemini-1.5-Pro & Retrieval & 42.5 & 19.8 & 24.0 & 39.3 & 27.4 & 41.0 & 31.6 \\
\bottomrule
\end{tabular}}
\caption{Results on SummHay using Gemini 1.5 Flash and Pro.}
\label{tab:summhay_gemini}
\end{table*}

%% file: tables/ablation_llama3.tex
\begin{table*}
\centering
\resizebox{\textwidth}{!}{
\begin{tabular}{@{}llrrrrrrr@{}}
\toprule
& & & & & & \multicolumn{3}{c}{A3CU} \\
Transformer & Method & ROUGE-1 & ROUGE-2 & ROUGE-L & ROUGE-Lsum & Recall & Precision & F1 \\
\midrule
\multicolumn{9}{c}{SummHay} \\
\midrule
Llama-3-8B & Hierarchical & 22.0 & 8.3 & 13.0 & 20.3 & 10.8 & 23.2 & 13.6 \\
Llama-3-8B & Incremental & 32.6 & 15.0 & 20.0 & 30.0 & 18.3 & 36.2 & 23.2 \\
Llama-3-70B & Hierarchical & 17.6 & 5.0 & 11.0 & 16.0 & 7.4 & 14.3 & 9.2 \\
Llama-3-70B & Incremental & 34.6 & 13.8 & 19.8 & 31.5 & 16.7 & 30.5 & 20.3 \\
\midrule
\multicolumn{9}{c}{SummHay (oracle)} \\
\midrule
Llama-3-8B & Hierarchical & 34.0 & 16.3 & 19.4 & 31.4 & 21.0 & 35.5 & 24.6 \\
Llama-3-8B & Incremental & 39.2 & 19.7 & 23.5 & 36.3 & 25.2 & 45.5 & 29.9 \\
Llama-3-70B & Hierarchical & 30.0 & 13.3 & 17.0 & 27.8 & 17.0 & 29.0 & 19.9 \\
Llama-3-70B & Incremental & 39.9 & 19.0 & 23.5 & 36.7 & 24.1 & 42.7 & 29.3 \\
\bottomrule
\end{tabular}}
\caption{Results on SummHay using the short context Llama-3 models.}
\label{tab:summhay_llama3}
\end{table*}

%% file: tables/ablation_chunk_size.tex
\begin{table*}
\centering
\resizebox{\textwidth}{!}{
\begin{tabular}{@{}llrrrrrrr@{}}
\toprule
& & & & & & \multicolumn{3}{c}{A3CU} \\
Transformer & Method & ROUGE-1 & ROUGE-2 & ROUGE-L & ROUGE-Lsum & Recall & Precision & F1 \\
\midrule
Llama-3.1-8B & Hierarchical-8K & 27.3 & 10.1 & 15.3 & 25.1 & 14.0 & 22.9 & 15.6 \\
Llama-3.1-8B & Hierarchical-16K & 30.8 & 12.6 & 17.6 & 28.4 & 16.7 & 27.9 & 18.9 \\
Llama-3.1-8B & Hierarchical-32K & 28.9 & 11.4 & 16.4 & 26.8 & 15.8 & 26.0 & 17.5 \\
\midrule
Jamba-1.5-Mini & Hierarchical-8K & 38.2 & 11.8 & 19.5 & 35.2 & 14.5 & 18.4 & 15.2 \\
Jamba-1.5-Mini & Hierarchical-16K & 37.7 & 12.0 & 20.4 & 34.5 & 14.7 & 19.9 & 16.0 \\
Jamba-1.5-Mini & Hierarchical-32K & 37.0 & 12.3 & 19.7 & 33.6 & 14.8 & 21.6 & 16.3 \\
\bottomrule
\end{tabular}}
\caption{Results on SummHay using different chunk sizes for the hierarchical method.}
\label{tab:summhay_chunk_size}
\end{table*}

%% file: tables/ablation_retriever.tex
\begin{table*}
\centering
\resizebox{\textwidth}{!}{
\begin{tabular}{@{}llrrrrrrr@{}}
\toprule
& & & & & & \multicolumn{3}{c}{A3CU} \\
Transformer & Method & ROUGE-1 & ROUGE-2 & ROUGE-L & ROUGE-Lsum & Recall & Precision & F1 \\
\midrule
Llama-3.1-8B & Retrieval-E5 & 50.1 & 25.1 & 28.6 & 47.3 & 33.9 & 35.1 & 33.2 \\
Llama-3.1-70B & Retrieval-E5 & 49.8 & 25.7 & 28.7 & 46.8 & 32.2 & 41.1 & 34.6 \\
Command-R & Retrieval-E5 & 44.8 & 19.3 & 24.5 & 41.5 & 27.2 & 36.7 & 29.5 \\
Jamba-1.5-Mini & Retrieval-E5 & 44.1 & 20.8 & 25.5 & 40.7 & 26.9 & 42.0 & 31.5 \\
\bottomrule
\end{tabular}}
\caption{Results on SummHay using the E5 RoPE retriever instead of SFR-Embedding-2.}
\label{tab:summhay_e5}
\end{table*}

%% file: tables/ablation_jamba.tex
\begin{table*}
\centering
\resizebox{\textwidth}{!}{
\begin{tabular}{@{}llrrrrrrr@{}}
\toprule
& & & & & & \multicolumn{3}{c}{A3CU} \\
Transformer & Method & ROUGE-1 & ROUGE-2 & ROUGE-L & ROUGE-Lsum & Recall & Precision & F1 \\
\midrule
Jamba-1.5-Mini-Grounded & Full-Context & 45.4 & 22.0 & 26.7 & 42.6 & 26.0 & 43.5 & 31.3 \\
Jamba-1.5-Mini-Grounded & Hierarchical & 40.8 & 14.3 & 20.9 & 37.7 & 18.6 & 20.2 & 18.2 \\
Jamba-1.5-Mini-Grounded & Incremental & 41.8 & 17.6 & 22.7 & 38.5 & 23.9 & 32.6 & 25.9 \\
Jamba-1.5-Mini-Grounded & Retrieval & 44.4 & 21.8 & 26.3 & 41.2 & 26.5 & 43.2 & 31.1 \\
\bottomrule
\end{tabular}}
\caption{Results on SummHay using grounded generation chat template with Jamba-1.5-Mini.}
\label{tab:summhay_jamba_grounded}
\end{table*}

%% file: tables/qual_hierarchical.tex
\begin{table*}
    \centering
    \resizebox{\textwidth}{!}{
    \begin{tabular}{p{24cm}}
    \toprule
    \textbf{Gold}: China announces its intent to increase tariffs from 10\% to 25\% on \$60 billion worth of imports from the United States effective June 1, 2019. \\
    \textbf{Prediction (best)}: US markets plummeted after China announced retaliatory tariffs on \$60 billion of US imports, including batteries, spinach, and coffee, in response to the Trump administration's hike on Chinese imports from 10\% to 25\%. \\
    \textbf{Prediction (final)}: The US and China engage in an escalating trade war, imposing tariffs on each other's goods, causing market losses, economic uncertainty, and affecting global economic growth, trade, and markets, with potential impacts on US consumers and businesses. \\
    \midrule
    \textbf{Gold}: Venezuela's Ambassador to Italy, Isaías Rodríguez, resigns, citing the government's lack of finances amid international sanctions on the Nicolás Maduro government. Rodríguez says, due to the sanctions, he has been unable to pay his staff or the rent for his office in Rome.\\
    \textbf{Prediction (best)}: The Venezuelan ambassador to Italy, Isaías Rodríguez, has resigned due to the country's financial difficulties caused by US sanctions. He cites inability to pay employees and rent, and will dedicate himself to being a grandfather, stating he ``cannot turn water into wine'' \\
    \textbf{Prediction (final)}: The document discusses Venezuela's humanitarian crisis, disputed re-election, and diplomatic efforts to resolve the crisis, including US sanctions, opposition negotiations, and international support for a democratic transition and potential regime change. \\
    \midrule
    \textbf{Gold}: One colleague mentions that their department uses specific productivity metrics like the number of tasks completed per week to measure the impact of remote work on employee productivity. A colleague argues that remote work results in fewer distractions compared to an open office environment, citing a 2019 study that found remote workers are 35\% less likely to be distracted by co-workers. Another colleague points out that remote employees often work more overtime, referencing a 2022 company internal report showing a 20\% increase in overtime hours since the shift to remote work. A colleague highlights that remote work allows for extended deep work periods, mentioning that they've personally experienced 3-hour uninterrupted work blocks at home, which were rarely possible in the office. One colleague mentions that remote work can align better with individual employees' peak productivity times, citing a 2020 survey where 60\% of employees reported higher productivity outside the traditional 9-to-5 work hours. \\
    \textbf{Prediction (best)}: Yes, productivity level change is discussed in the conversation. \textbackslash n\textbackslash n Colleague 3 mentions a 2019 study that found remote workers are 35\% less likely to be distracted by co-workers compared to those in an open office, suggesting that remote work can be a productivity booster. Colleague 1 also agrees that the solitude of remote work can lead to deep focus and increased productivity. \textbackslash n\textbackslash n Additionally, colleague 2 compares the office environment to a marching band parading through the office, emphasizing how the constant interruptions can break one's flow and hinder productivity. This highlights the potential benefits of remote work in terms of increased productivity due to fewer distractions. \textbackslash n\textbackslash n Overall, the conversation suggests that remote work can have a positive impact on productivity, with fewer distractions and interruptions allowing employees to focus more effectively. \\
    \textbf{Prediction (final)}: Yes, productivity level change is discussed. \\
    \midrule
    \textbf{Gold}: Suspected Baloch nationalist gunmen storm a coal mine in Islamabad, Pakistan, killing at least five people. \\
    \textbf{Prediction (best)}: At least five people were killed in a gun and bomb attack on a coal mine in southwestern Pakistan by ethnic Baloch separatists, with two miners, two security personnel, and a driver among the dead, amid escalating violence in the region. \\
    \textbf{Prediction (final)}: The document reports various news articles from 2019, including attacks on hotels in Gwadar, Pakistan, by the Balochistan Liberation Army, as well as human trafficking and security concerns during the Muslim fasting month of Ramadan. \\
    \midrule
    \bottomrule
    \end{tabular}}
    \caption{Examples illustrating loss of salient information in Hierarchical method.}
    \label{tab:qual_hierarchical}
\end{table*}

%% file: tables/qual_incremental.tex
\begin{table*}
    \resizebox{\textwidth}{!}{
    \begin{tabular}{p{24cm}}
    \toprule
    \textbf{Gold}: One student shares that they do 10 minutes of deep breathing exercises each night before going to bed to help reduce stress. A student mentions that taking a 30-minute walk each afternoon helps clear their mind and manage stress levels during exam preparation. One of the students suggests taking a 5-minute break after every 25 minutes of studying, and mentions the Pomodoro technique as a helpful method. A student recommends using a specific meditation app called 'Calm' that they use for 15 minutes each morning to manage stress. Another student shares their habit of writing down 3 positive affirmations about their capabilities each day, which they find significantly helps reduce their stress. \\
    \textbf{Prediction (best)}: Yes, the students indirectly shared exam stress tips through their conversation. They discussed various techniques and strategies that helped them manage stress and stay focused during exam preparation, including:\textbackslash n\textbackslash n* Maintaining a consistent sleep schedule and taking breaks\textbackslash n* Using the Pomodoro Technique to avoid burnout\textbackslash n* Incorporating digital detox into their daily routine\textbackslash n* Practicing deep breathing exercises\textbackslash n* Taking short walks to clear their minds\textbackslash n* Trying relaxation techniques such as meditation\textbackslash n* Using spaced repetition with the Anki app\textbackslash n* Creating mind maps for visual learners\textbackslash n* Using a meditation app like 'Calm'\textbackslash n\textbackslash n They also shared specific experiences and recommendations, such as Alejandro's 30-minute afternoon walk, which helped him manage stress and clear his mind. They also discussed the potential benefits of physical activity, meditation, and relaxation techniques in reducing anxiety and improving focus. The conversation was supportive and encouraging, with the students offering to help and motivate each other throughout the exam preparation period. \\
    \textbf{Prediction (final)}: No, they did not directly share exam stress tips. Their conversations centered on sharing personal experiences and strategies related to studying, time management, prioritization, and productivity, which indirectly contributed to managing exam stress. They discussed various approaches to studying, adapting their study plan, and organizing information, but did not explicitly mention stress management techniques like deep breathing exercises or the Pomodoro technique. Their focus was on developing a study plan, sharing notes, and reviewing material together to prepare for the exam, rather than directly addressing stress management. \\
    \midrule
    \textbf{Gold}: Prince Harry and Meghan, the Duke and Duchess of Sussex, name their son Archie Harrison Mountbatten-Windsor. \\
    \textbf{Prediction (best)}: Prince Harry and Meghan, Duchess of Sussex, have named their baby boy Archie Harrison Mountbatten-Windsor. The palace announced the name after the couple posed with the two-day-old for the media, meeting Queen Elizabeth II for the first time. \\
    \textbf{Prediction (final)}: Prince Harry and Meghan Markle introduced their son Archie to the world at Windsor Castle, presenting a more casual and intimate image, contrasting with traditional royal protocol. They shared parenting duties and responsibilities, showcasing a new generation of feminist royals.\\
    \midrule
    \textbf{Gold}: The United States lifts tariffs on steel and aluminum from Canada and Mexico. \\
    \textbf{Prediction (best)}: The US will lift tariffs on steel and aluminum from Canada and Mexico, resolving a yearlong deadlock and providing economic relief to Detroit automakers, who paid billions in increased duties, and addressing a key obstacle to ratifying the new North American trade pact. \\
    \textbf{Prediction (final)}: The number of asylum seekers in Canada tripled between 2015 and 2018, with a peak of 8,801 claims in August 2017. Most claims came from Haitians, Nigerians, and US citizens, with many fleeing the US due \\
    \midrule
    \bottomrule
    \end{tabular}}
    \caption{Examples illustrating loss of salient information in Incremental method.}
    \label{tab:qual_incremental}
\end{table*}